\documentclass[10pt,twocolumn,letterpaper]{article}
\pdfoutput=1
\usepackage{cvpr}
\usepackage{pifont}
\usepackage{times}
\usepackage{epsfig}
\usepackage{paralist}
\usepackage{graphicx}
\usepackage{amsmath}
\usepackage{amsfonts}
\usepackage{xspace}
\usepackage{amssymb}
\usepackage{xcolor}
\usepackage{wrapfig,lipsum,booktabs}
\usepackage{cuted}
\usepackage{capt-of}
\usepackage[percent]{overpic}

\usepackage{subcaption}
\newcommand{\name}{Im2Vec\xspace}

\usepackage[pagebackref=true,breaklinks=true,colorlinks,bookmarks=false]{hyperref}

\cvprfinalcopy 

\def\cvprPaperID{16} 

\newcommand{\notation}[1]{\ensuremath{#1}\xspace}

\newcommand{\RasterImage}{\notation{I}}
\newcommand{\RasterOutput}{\notation{O}}
\newcommand{\RasterLayer}[1]{\notation{L_{#1}}}
\newcommand{\LatentCode}{\notation{z}}
\newcommand{\LatentDimension}{\notation{d}}
\newcommand{\PathLatentCode}[1]{\notation{z_{#1}}}
\newcommand{\PathDepth}[1]{\notation{d_{#1}}}
\newcommand{\NumPyrLevels}{\notation{L}}

\newcommand{\PositionCircle}{\notation{p}}
\newcommand{\CircleOffset}{\notation{\delta p}}
\newcommand{\ControlPointPosition}{\notation{x}}
\newcommand{\PointType}{\notation{c}}
\newcommand{\NumSegments}{\notation{k}}
\newcommand{\NumSamples}{\notation{3\NumSegments}}
\newcommand{\NumShapes}{\notation{T}}

\newcommand{\Emojis}{\textsc{Emojis}\xspace}
\newcommand{\Icons}{\textsc{Icons}\xspace}
\newcommand{\MNIST}{\uppercase{mnist}\xspace}
\newcommand{\Fonts}{\textsc{Fonts}\xspace}

\ifcvprfinal\fi
\begin{document}

\title{Im2Vec: Synthesizing Vector Graphics without Vector Supervision}
\author{Pradyumna Reddy$^{1}$
\and
Micha\"{e}l Gharbi$^{2}$
\and
Michal Luk\'{a}\v{c}$^{2}$
\and
Niloy J. Mitra$^{1,2}$\smallbreak
\and
$^{1}$University College London \quad
$^{2}$Adobe Research
}
\maketitle

\begin{abstract}
    Vector graphics are widely used to represent fonts, logos, digital
    artworks, and graphic designs.
    But, while a vast body of work has focused on generative algorithms for
    raster images, only a handful of options exists for vector graphics.
    One can always rasterize the input graphic and resort to image-based
    generative approaches, but this negates the advantages of the
    vector representation.
    The current alternative is to use specialized models that require explicit
    supervision on the vector graphics representation at training time.
    %
    This is not ideal because large-scale high-quality vector-graphics datasets
    are difficult to obtain.
    Furthermore, the vector representation for a given design is not unique, so
    models that supervise on the vector representation are unnecessarily
    constrained.
    Instead, we propose a new neural network that can generate complex vector
    graphics with varying topologies, and only requires indirect supervision
    from readily-available \emph{raster} training images (i.e., with no vector
    counterparts).
    To enable this, we use a differentiable rasterization pipeline that renders
    the generated vector shapes and composites them together onto a raster
    canvas.
    %
    %
    We demonstrate our method on a range of datasets, and provide comparison
    with state-of-the-art SVG-VAE and DeepSVG, both of which require explicit
    vector graphics supervision.
    Finally, we also demonstrate our approach on the MNIST dataset, for which no
    groundtruth vector representation is available. 
    Source code, datasets and more results are available at \url{http://geometry.cs.ucl.ac.uk/projects/2021/Im2Vec/}.
\end{abstract}

\section{Introduction}

In vector graphics, images are represented as collections of parametrised shape primitives rather than a regular raster of pixel values. This makes for a compact, infinitely scalable representation with appearance that may be varied at need simply by modifying stroke or colour parameters. As a result, it is favoured by graphic artists and designers.


\begin{figure}
    \centering
    \includegraphics[width=\columnwidth]{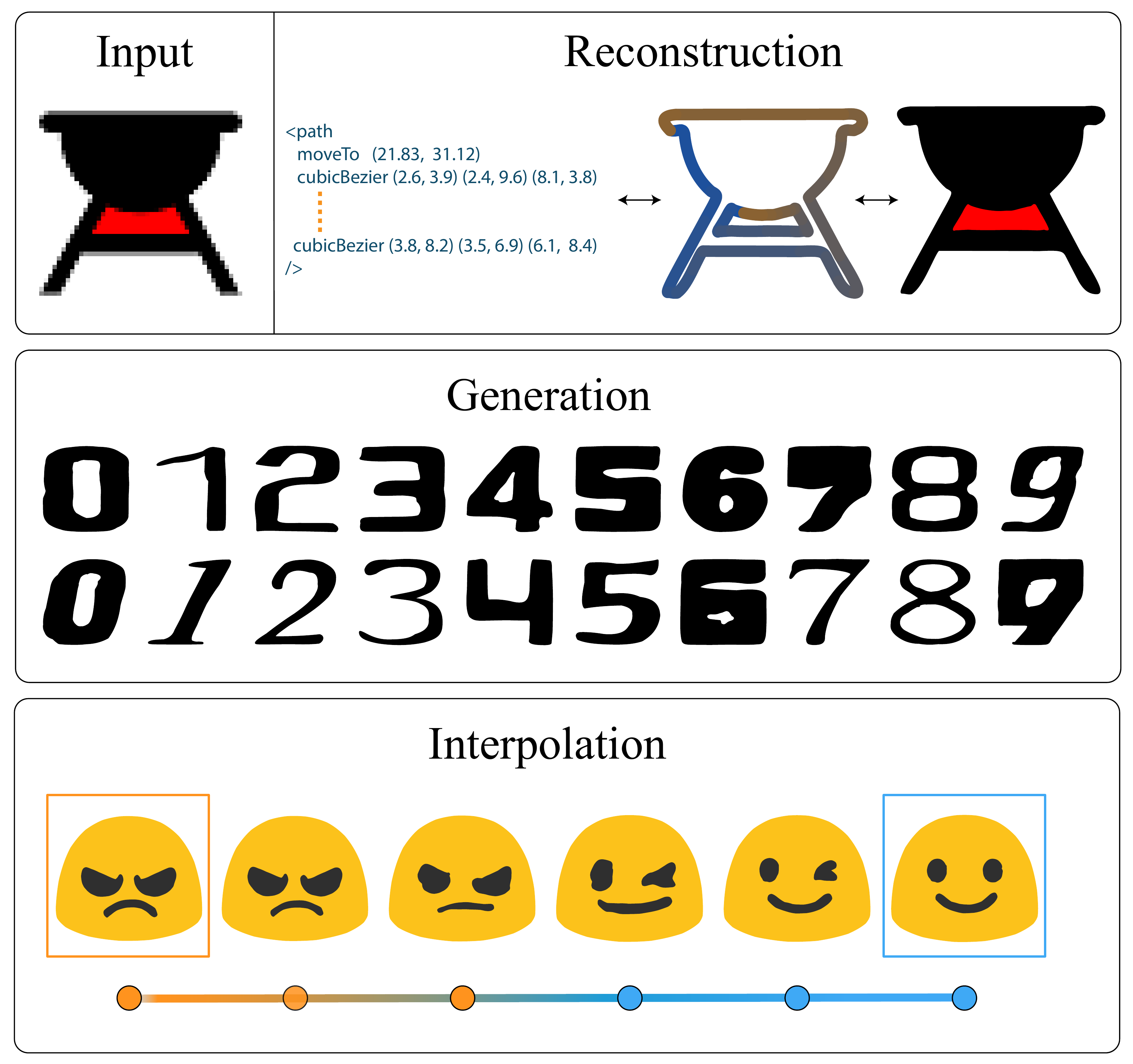}
    \caption{We present \name that can be trained with only image supervision  to produce a latent space for vector graphics output. The learned space supports reprojection, sampling (i.e., generation), and interpolation. }
    \label{fig:teaser}
\end{figure}

Unfortunately, creating vector graphics still remains a difficult task largely limited to manual expert workflows, because the same irregular structure makes it ill-suited for today's convolution-based generative neural architectures. There is demand for a generative approach suitable for this domain, but it is not yet well served by research because of the difficult design requirements. Suitable approaches should:
%
\begin{inparaenum}[(i)]
\item produce output in vector format;
\item establish correspondence across elements of the same family;
\item support reconstruction, sampling, and interpolation;
\item give user control over accuracy versus compactness of the representation;
    and finally,
\item be trainable directly using images without the need for vector supervision. 
\end{inparaenum}

SVG-VAE~\cite{lopes2019learned} and DeepSVG~\cite{carlier2020deepsvg}, the two
leading generative algorithms for vector graphics, cast synthesis as a sequence
prediction problem, where the graphic is a sequence of drawing instructions,
mimicking how common formats actually represent vector art.
Training these methods therefore requires supervision from ground truth vector
graphics sequences, which are difficult to collect in large volumes.
Furthermore, the mapping from sequences of parametrised drawing instruction to
actual images is highly non-linear with respect to the parameters and also
non-injective, allowing a variety of different sequences to produce the same
visual result. This makes it difficult to consider appearance as a criterion,
and also causes the produced results to inherit any structural bias baked into
the training sequences.
    
An approach aiming to do away with such vector supervision would need to
overcome a number of challenges. First, the relationship between the
representation and its appearance must be made explicit and differentiable.
Second, it must operate on an internal representation that directly maps to a
vector graphics representation and is flexible enough to support a large range
of topologies and shape complexities. Finally, it should extract correspondences
between related shapes, directly from unlabelled images. 

\begin{figure}[h!]
    \centering
    \begin{overpic}[width=\linewidth,tics=5]{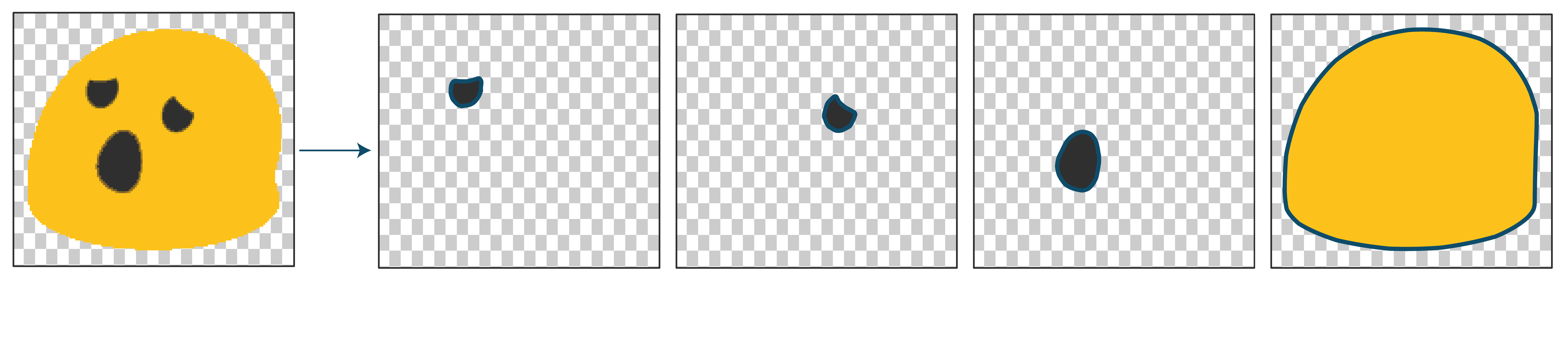}
        \put (9, 1) {\RasterImage}
        \put (29, 1) {\RasterLayer{1}, \PathDepth{1}}
        \put (47, 1) {\RasterLayer{2}, \PathDepth{2}}
        \put (65, 1) {\RasterLayer{3}, \PathDepth{3}}
        \put (78.5, 2) {\ldots}
        \put (85, 1) {\RasterLayer{\NumShapes}, \PathDepth{\NumShapes}}
    \end{overpic}
    \caption{\label{fig:illustration_layering}
        \name encodes a shape as a layered set of filled curves (or shapes).
        Each shape is obtained by deformation of a topological disk,
        differentiably rasterized into images \RasterLayer{i}, then
        differentiably composited back-to-front according to scalar depth
        variables \PathDepth{i}. 
    }
\end{figure}
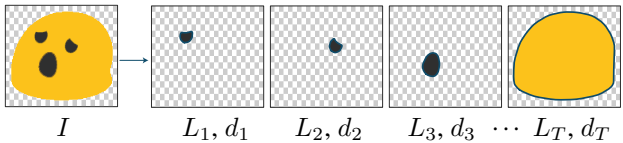

In this paper, we propose such a method, called \name, based on a representation
that mimics the compositing behaviour of complex vector graphics. It uses a
variable-complexity closed B\'{e}zier path as the fundamental primitive, with
the capability to composite a variable number of these to create shapes of
arbitrary complexity and topology (shown in
Figure~\ref{fig:illustration_layering}).

The key insight that allows the handling of arbitrary complexity is that we can
treat any primitive closed shape as a deformation of a unit circle, which is
modelled as 1D convolution on samples from this circle conditioned on a common
latent vector. By recombining these primitive paths through a differentiable
rasterizer~\cite{li2020diffvg} and differentiable
compositing~\cite{reddy2020diffcompose}, we can natively represent vector art
while learning to generate it purely based on appearance, obviating the need for
vector supervision.

We evaluate \name on a variety of examples with varying complexity and topology
including fonts, emojis, and icons.
We demonstrate that \name, even without any vector supervision, consistently
performs better reconstruction compared to SVG-VAE and DeepSVG when trained on
the same dataset.
We also compare our approach to a purely raster-based autoencoder, which we dub
ImageVAE.
While ImageVAE and \name produce comparable reconstruction quality, \name
outputs vector graphics and hence enjoys the associated editability and
compactness benefits.
Finally, we quantify the compactness versus approximation power of our method,
and demonstrate \name can be used to vectorize the MNIST dataset for which no
groundtruth vector representation is available. 

\section{Related Work}
Deep learning techniques for parametric vector shapes have recently garnered
significant interest from the machine learning community~\cite{jayaraman2020uv, gao2019deepspline, groueix2018papier, zheng2018strokenet, paschalidou2019superquadrics}.

\paragraph{Learning-based image vectorization.}
Our autoencoder encodes raster images. It can therefore address the single-image
vectorization problem \cite{bessmeltsev2019vectorization, favreau2016fidelity, selinger2003potrace, kopf2011depixelizing, dominici2020vectorization, hoshyari2018vectorization},
 for which learning-based solutions have been proposed.
Egiazarian~et~al.~\cite{egiazarian2020deep} vectorize technical line drawings.
They predict the parameters of vector primitives using a transformer-based
network, and refine them by optimization.
DeepSpline~\cite{gao2019deepspline} produces parametric curves of variable
lengths from images using a pre-trained VGG network~\cite{Simonyan15} for
feature extraction followed by a hierarchical recurrent network.
Guo~et~al.~\cite{guo2019deep} use neural networks sub-divide line drawings and
reconstruct the local topology at line junctions. The network predictions are
used in a least squares curve fitting step to estimate B\'ezier curve parameters.
Liu~et~al.~\cite{liu2017raster} focus on vectorization of rasterized floorplans.
They use a network to extract and label wall junctions, and use this information
to solve an integer program that outputs the vectorized floor plans as a set of
architectural primitives.
These works produce high-quality vectorizations but, unlike ours, focus on
the single image case.
In contrast, our objective is to train a latent representation which can serve
both for vectorization of existing raster images, and for generating new
graphics by sampling with no post-processing.

\paragraph{Parametric shape estimation.}
Deep learning methods for parametric shape estimation typically encode
shapes as an assembly of primitives, often with fixed topology and
cardinality~\cite{groueix2018papier}.
Smirnov~et~al.~\cite{smirnov2020deep} fit rasterized fonts using quadratic
B\'ezier curves, and 3D signed distance fields using cuboids. 
Their outputs have predetermined, fixed topologies that are specified as
class-dependent templates.
Zou~et~al.~\cite{zou20173d} train a recurrent network that predict shapes as a
collection of cuboids from depth maps; they supervise directly on the shape parameters.
Tulsiani~et~al~\cite{tulsiani2017learning} also use hierarchies of cuboids, but
from occupancy volumes.
Similar techniques have explored other primitives like 
superquadrics~\cite{paschalidou2019superquadrics} and Coon
patches~\cite{smirnov2019deep} as
primitives.
Sinha~et~al.~\cite{sinha2017surfnet} represents watertight 3D shapes as
continuous deformation of a sphere. This is analogous to our representation of
closed 2D curves.

\paragraph{Shape-generating programs.}

Ganin~et~al.~\cite{ganin2018synthesizing}, 
Huang~et~al.~\cite{huang2019learning}, and Nakano~\cite{nakano2019neural} train
Reinforcement Learning (RL) drawing agents.
They circumvent the need for direct supervision on the drawing program by
simulating a rendering engine to produce images from which they compute a reward
signal.
Ellis~et~al.~\cite{ellis2017learning} use program synthesis to generate graphics
expressed using a subset of the \LaTeX\xspace  language from hand drawings.
They do not work with complex parametric shapes like B\'ezier curves, which are
the basic building block of most vector designs.
Another notable work is the CSGNet~\cite{sharma2017csgnet} that present impressive performance in
estimating constructive solid geometry programs. It uses the REINFORCE~\cite{sutton2000policy} algorithm to learn in an
unsupervised manner, but runs into issues like drawing over previous predictions
in the later stages of the generation process.
Further, it can only output $32\times 32$ raster images, which lacks
the flexibility of vector graphics and is insufficient for applications
that require high fidelity.
Strokenet~\cite{zheng2018strokenet} trains an agent that draws strokes
after observing a canvas image and a generator that maps stroke parameters to a
new image.

\paragraph{Generative vector graphics model.}
Our goal is to obtain a generative model for vector graphics.
Previous works in this area have focused predominantly on the case where
direct vector supervision is available.
In contrast, our model can be trained from raster data alone.
SketchRNN~\cite{ha2017neural} introduces a model for both conditional and
unconditional sketch generation.
Sketches are encoded as a sequence of pen position and on/off states.
An LSTM is then trained to predict the parameters of a density function over the
sketch parameter space, which can then be sampled to produce a new sketches.
Similarly, Sketchformer~\cite{ribeiro2020sketchformer} proposed a
transformer based architecture  for encoding vector form sketches.
They show how the encoding can be used for sketch classification, image
retrieval, and interpolation.

SVG-VAE~\cite{lopes2019learned} is the first method that
attempts to estimate vector graphics parameters for generative tasks.
They follow a two stage training process.
First, they train an image Variational Auto Encoder (VAE).
Second, they freeze the VAE's weights and train a decoder that predicts
vector parameters from the latent variable learned on images.
They show a style-transfer application from one vector graphic to another.
Unlike ours, their method is not end-to-end, and it requires vector supervision.
More recently, 
DeepSVG~\cite{carlier2020deepsvg} showed that models operating on vector
graphics benefit from a hierarchical architecture; they demonstrate
interpolation and generation tasks.
Prior works~\cite{azadi2018multi,gao2019artistic} can generate new font glyphs
from partial observations, but they only work in a low-resolution
raster domain.
Li~et~al.~\cite{li2020diffvg} have recently proposed a
differentiable rasterizer that enables gradient based optimization and 
learning on vector graphics, using raster-based objectives.
This is a key building block for our method.
However, we go beyond the generative models they demonstrate.
In particular, our network can generate graphics made up of closed curves with
complex and varying topologies; it does not produce artifacts like overlapping
paths.

\section{Method}\label{sec:method}

\begin{figure*}[t!]
    \begin{overpic}[width=\linewidth]{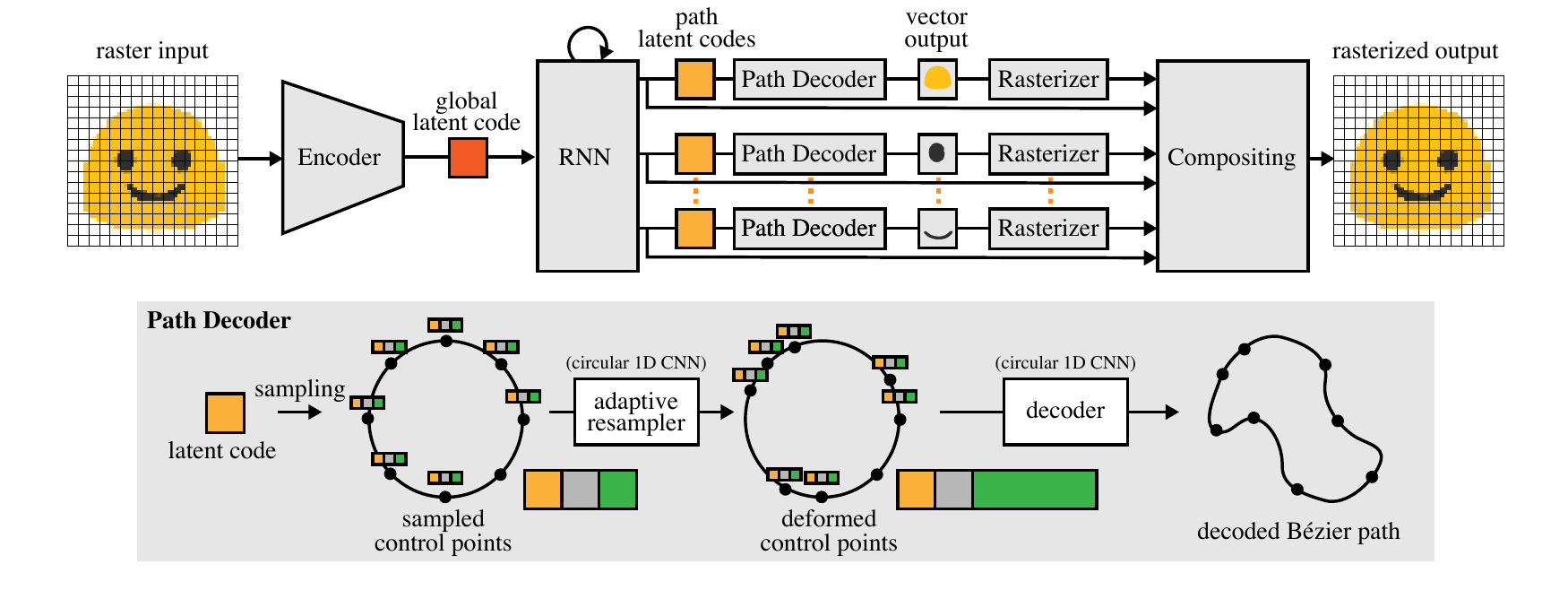}
        \put (9.5, 36.0) {\RasterImage}
        \put (89.5, 36.0) {\RasterOutput}
        \put (29.3, 27.8) {\LatentCode}

        \put (43.4, 32.7) {\PathLatentCode{1}}
        \put (43.4, 28) {\PathLatentCode{2}}
        \put (43.4, 23.3) {\PathLatentCode{\NumShapes}}
        \put (71, 33.7) {\scriptsize \RasterLayer{1}}
        \put (71, 31.7) {\scriptsize \PathDepth{1}}
        \put (71, 28.9) {\scriptsize \RasterLayer{2}}
        \put (71, 27.0) {\scriptsize \PathDepth{2}}
        \put (71, 24.4) {\scriptsize \RasterLayer{\NumShapes}}
        \put (71, 22.4) {\scriptsize \PathDepth{\NumShapes}}
        %
        %
        \put (13.7, 11.5) {\PathLatentCode{t}}
        \put (34.0, 6.5) {\PathLatentCode{t}}
        \put (36.3, 6.5) {$\PointType_i$}
        \put (38.6, 6.5) {$\PositionCircle_i$}
        \put (57.7, 6.5) {\PathLatentCode{t}}
        \put (60.1, 6.5) {$\PointType_i$}
        \put (62.4, 6.5) {$\PositionCircle_i + \CircleOffset_i$}
        \put (86.1, 12) {$\ControlPointPosition_i$}
    \end{overpic}
    \caption{\label{fig:pipeline}
        {\bf Architecture overview.}
        We train an end-to-end variational autoencoder that encodes a raster image
        to a latent code \LatentCode, which is then decoded
        to a set of ordered closed vector paths ({\bf top}).
        We then rasterize the paths using DiffVG~\protect{\cite{li2020diffvg}}
        and composite them together using DiffComp to obtain a rasterized output,
        which we compare to the ground truth raster target for supervision at 
        training time.
        Our model can handle graphics with multiple component paths.
        It uses an RNN to produce a latent code \PathLatentCode{t} for each path, from the global
        latent code \LatentCode representing the graphic as a whole.
        Our path decoder ({\bf bottom}) decodes the path codes into closed
        B\'ezier paths.
        Our representation ensures the paths are closed by sampling the path
        control points uniformly on the unit circle.
        These control positions are then deformed using a 1D convolutional
        network with circular boundary conditions to enable adaptive control
        over the point density.
        Finally, another 1D circular CNN processes the adjusted points on the
        circle to output the final path control points in the absolute
        coordinate system of the drawing canvas.
        The auxiliary network that predicts the optimal number of control points
        per path is trained independently from our main model; it is not shown
        here.
    }
\end{figure*}
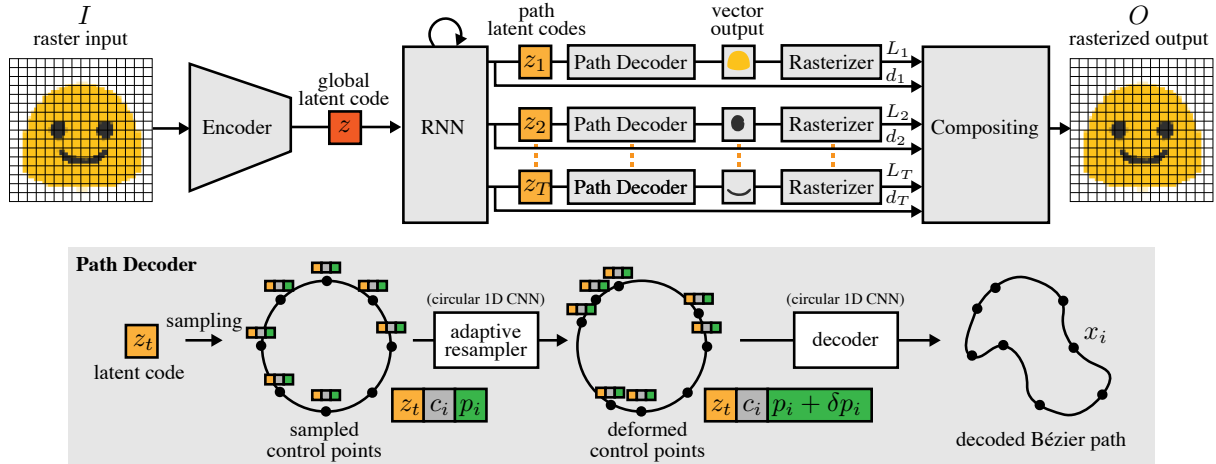

Our goal is to build a generative model for vector graphics that does not
require vector supervision, i.e., that only requires raster images at training
time.
Our model follows an encoder--decoder architecture (Fig.~\ref{fig:pipeline}).
The encoder has a standard design~\cite{he2015deep}; it maps a raster image
\RasterImage to a latent variable $\LatentCode\in\mathbb{R}^\LatentDimension$,
which is then decoded into a vector graphic structure.
Our decoder has been carefully designed so that it can generate complex
graphics, made of a variable number 
\NumShapes of paths, with varying lengths and no predetermined
topology (\S~\ref{sec:decoder}).
We also train an auxiliary model to predict the optimal number of control points
for each path (\S~\ref{sec:complexity_predictor}).
Finally, each vector shape is rasterized using a differentiable
rasterizer~\cite{li2020diffvg} and composited into a final rendering
~\cite{reddy2020diffcompose}, which we compare to a raster
ground truth for training (\S~\ref{sec:multires_loss}).

\subsection{Vector Graphics Decoder}\label{sec:decoder}

We choose to represent a vector graphic as a depth-ordered set of \NumShapes
closed B\'ezier paths, or equivalently, a set of \NumShapes simply connected
solid 2D shapes.
The first operator in our decoder is a recurrent neural network (RNN) that
consumes the global latent code \LatentCode representing the graphic as a whole
(\S~\ref{sec:rnn}).
At each time step $t$, the RNN outputs a per-path latent code $\PathLatentCode{t}$.
This mechanism lets us generate graphics with arbitrary numbers of paths,
and arbitrary topology (using fill rules to combine the shapes).
The path-specific codes are then individually processed by a \emph{path decoder}
module (\S~\ref{sec:path_decoder}) which outputs the parameters of a closed
path of arbitrary length using cubic B\'ezier segments.

\subsubsection{Single path decoder with circular convolutions}\label{sec:path_decoder}
To ensure the individual paths are closed, we obtain them by continuous deformation of the
unit circle.
Specifically, for each shape, we sample \NumSamples points along the circle,
corresponding to the control points of $\NumSegments$ cubic B\'{e}zier
segments. 
We compute the 2D cartesian coordinates $\PositionCircle_i$ of each of these points, and
annotate them with a 1-hot binary variable $\PointType_i$ to distinguish between
the segment endpoints --- every third point, which the B\'{e}zier path
interpolates --- and the other control points.

We replicate the path's latent code $\PathLatentCode{t}$ and concatenate it with
the sample position and point type label, so that each sample on the circle is
represented as a vector $\begin{bmatrix}\PositionCircle_i & \PointType_i &
\PathLatentCode{t}\end{bmatrix}$, $i\in\{1,\ldots,\NumSamples\}$,
which we call a \emph{fused latent vector}.
These are then arranged into a cyclic buffer, which is then processed by a
neural network performing 1D convolutions with cyclic boundary conditions (along
the sample dimension) to obtain the final spatial locations of the path's
control points:
$\ControlPointPosition_1,\ldots,\ControlPointPosition_{\NumSamples}$.
The cyclic convolution along the sample axis corresponds to
convolution along the perimeter of the unit circle.
It is a crucial component of our method because it enables information sharing
between neighbouring samples, while respecting the closed topology of the shape.
We use 3-tap filters for all convolutions and ReLU activations.

\begin{figure}[tb!]
    \centering
    \begin{subfigure}[b]{\linewidth}
        \centering
        \includegraphics[width=\linewidth]{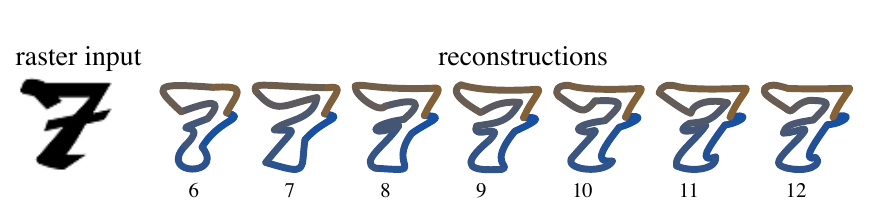}
        \caption{\label{fig:error_vs_segments}visual fidelity vs.\ number of
        segments}
    \end{subfigure}
    \begin{subfigure}[b]{\linewidth}
        \centering
        \includegraphics[width=\linewidth]{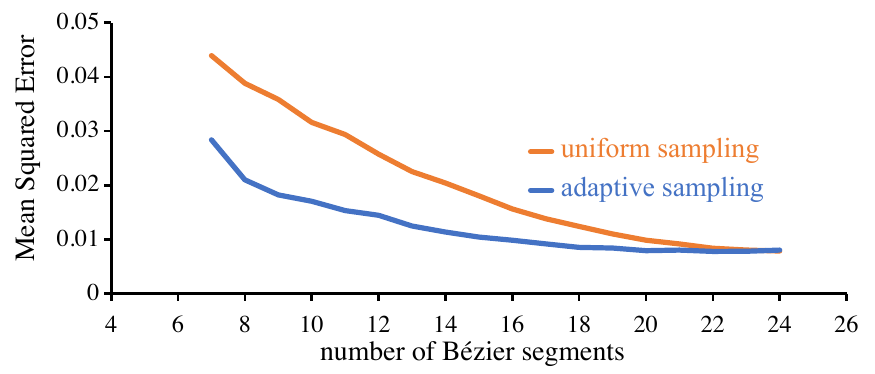}
        \caption{error vs.\ number of segments\label{fig:adaptive}}
    \end{subfigure}
    \caption{\label{fig:sample_deform_graph}
    {\bf Uniform vs.\ adaptive sampling.}
    Our decoder provides a natural control over the complexity of the vector
    graphics it produces.
    By adjusting the sampling density on the unit circle, we can increase the
    number of B\'ezier segments and obtain a finer or vector representation of a
    target raster image \subref{fig:error_vs_segments}.
    Our adaptive sampling mechanism (\S~\ref{sec:sampling}) improves
    reconstruction accuracy, compared to a uniform distribution of the
    control points with the same number of segments \subref{fig:adaptive}.
    This adaptive scheme achieves good reconstructions with as few as 7--8
    segments, while uniform sampling requires 12--14.
    }
\end{figure}

Sampling the unit circle rather than using a fixed-length input array allows us
to adjust the complexity (i.e., the number of segments $\NumSegments$) of the
B\'ezier path by simply changing the sampling density.
In Section~\ref{sec:complexity_predictor}, we show this sampling density can be
determined automatically, based on complexity of the shape to match, using an
auxiliary network.
Figure~\ref{fig:sample_deform_graph} shows the impact of the number of segments
on the reconstruction quality.

\subsubsection{Adaptive control point density}\label{sec:sampling}

The most natural choice for our control point parameterization would be to choose 
equally spaced sample points along the unit circle (in angle).
We found this uniform control points allocation was often sub-optimal.
Ideally, more control points should be allocated to sections of the path with
higher complexity (e.g., sharp creases or serifs for fonts).
To address this, we propose an adaptive sampling mechanism, which we call the
\emph{sample deformation} subnetwork.
This module is a 1D convolutional network with cyclic boundary condition acting
on the fused latent vectors 
$\begin{bmatrix}\PositionCircle_i & \PointType_i & \PathLatentCode{t}\end{bmatrix}$,
where the $\PositionCircle_i$ are uniformly spaced along the circle.
It outputs a displacement $\CircleOffset_i$ for each sample point. We
parameterize this output in polar coordinates so that
$\PositionCircle_i + \CircleOffset_i$ remains on the circle. 

With our adaptive sampling mechanism turned on, the path decoder now operates on
the fused latent vector with sample deformation, $\begin{bmatrix}\PositionCircle_i + \CircleOffset_i & \PointType_i & \PathLatentCode{t}\end{bmatrix}$, instead of the
regularly-spaced positions.
In Figure~\ref{fig:adaptive}, we show the sample deformation module
improves the reconstruction accuracy, especially when few segments are used.
The benefit over the uniform sampling distribution diminishes as more curve
segments are added.

\subsubsection{Decoding multi-part shapes using an RNN}\label{sec:rnn}
So far, we have discussed a decoder architecture for a single shape,
but our model can represent vector graphics made of multiple parts.
This is achieved using a bidirectional LSTM~\cite{schuster1997bidirectional}
that acts on the graphic's latent code \LatentCode.
To synthesize a graphic with multiple component shapes, we run the recurrent
network for \NumShapes steps, in order to obtain shape latent codes for each
shape: $\PathLatentCode{1},\ldots,\PathLatentCode{\NumShapes}$.
We set \NumShapes to a fixed value, computed before training, equal to the
maximum number of components a graphic in our training dataset can have.
When a graphic requires fewer than \NumShapes shapes, the extra paths
produced by the RNN are degenerate and collapse to a single point; we discard them before rendering.

In addition to the shape latent codes \PathLatentCode{i}, the recurrent network
outputs an unbounded scalar depth value \PathDepth{i} for each path which is used by our differentiable compositing module when
rasterizing the shapes onto the canvas.

\subsection{Predicting the number of path control points}\label{sec:complexity_predictor}

Each path (shape) in our vector output can be made of a variable number of
segments.
Figure~\ref{fig:error_vs_segments} shows how the reconstruction loss
decreases as we increase the number of curve segments from 6-25, for multiple
designs.
It also shows that, depending on the design's complexity, not all paths 
need many segments to be represented accurately.
We train an auxiliary network conditioned on a path latent variable \PathLatentCode{t}
to model the complexity--fidelity trade-off and automatically determine the
optimal number of segments for a path.
This auxiliary network has 3 fully connected layers.
It outputs 3 parameters $a$, $b$, and $c$ of a parametric curve $x
\mapsto ae^{-bx} +c$ that approximates the loss graph of a given
shape, with respect to the number of segments.
Given this parametric approximation, we allow the user to set the quality
trade-off as a threshold on the derivative of the parametric curve.
Specifically, we solve for $x$ in the derivative expression and round up to
obtain the number of segments to sample.
This threshold defines what improvement in the reconstruction error is worth the
added complexity of an additional B\'ezier segment. Please refer to our supplementary for more information on the auxiliary network.

\begin{figure}[tb]
    \centering
    \includegraphics[width=\linewidth]{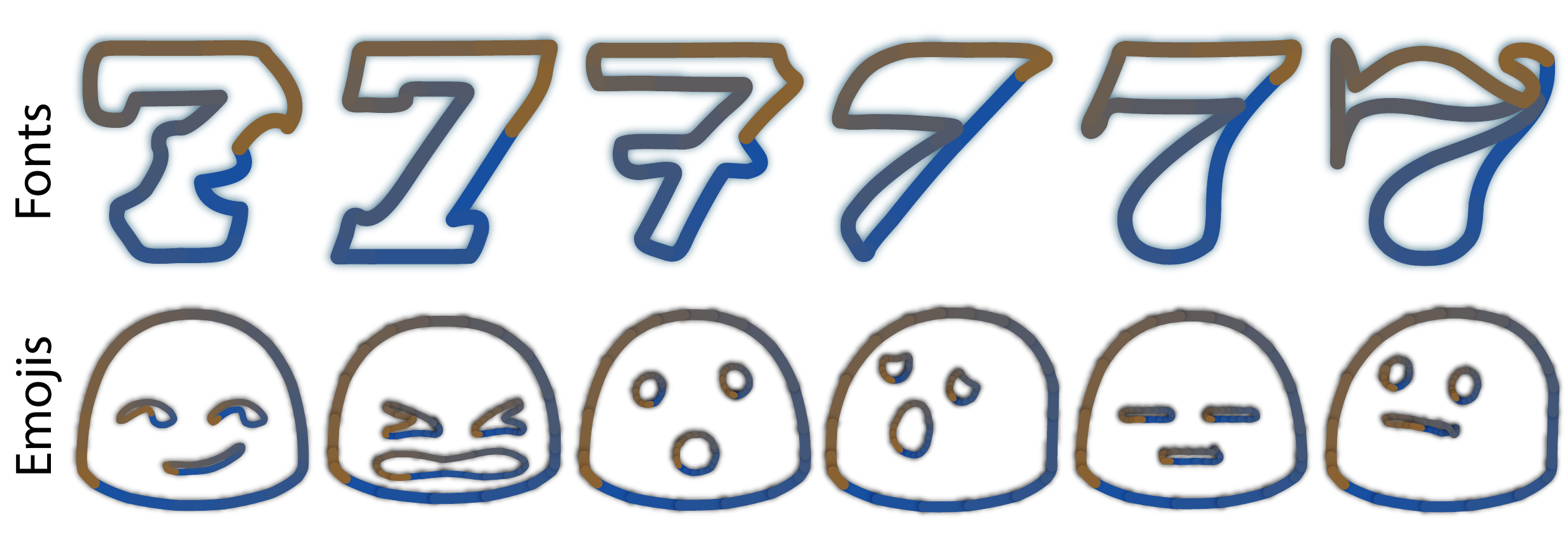}
    \caption{\label{fig:correspondence}
        {\bf Latent space correspondences.}
        \name encodes shapes as deformation of a topological
        disk.
        This naturally gives a point-to-point correspondence between shapes 
        across graphics design once we encode them in our latent space.
        Graphics can be made of a single path~(top), or multiple paths~(bottom). 
        In both cases, our model establish meaningful geometric correspondences
        between the designs, indicated by the blue--orange color coding.
    }
\end{figure}

\subsection{Multi-resolution raster loss}\label{sec:multires_loss}

Given a raster input image \RasterImage, our model encodes the design into a
global latent code \LatentCode, which the RNN decomposes into path latent codes
$\PathLatentCode{1},\ldots,\PathLatentCode{\NumShapes}$.
Our path decoder maps each path latent code to a closed B\'ezier path.
We rasterize each path individually, as a solid shape using the
differentiable rasterizer of Li~et~al.~\cite{li2020diffvg}, and composite them
together into a final raster image \RasterOutput using the differentiable compositing
algorithm of Reddy~et~al~\cite{reddy2020diffcompose}.
Since every step of the pipeline is differentiable, we can compute a loss
between input image \RasterImage and raseterized generated vector graphic \RasterOutput, and backpropagate the error to train our
model using gradient descent.

When we differentiate \RasterOutput with respect to the B\'ezier
parameters, the gradients have a small area of influence, corresponding to the
support of the rasterization prefiltering kernel.
This adversely affects convergence especially when the mismatch between
\RasterImage and \RasterOutput is high (e.g., at the early stages of the
training).
We alleviate this issue by rasterizing our graphics at multiple resolutions.
That is, we render an image pyramid instead of a single image, and aggregate the
loss at each pyramid level.
We obtain the ground truth supervision for each level by decomposing the target
image into a Gaussian pyramid, where each level is downsampled by a factor 2
along each dimension from the previous level.
The gradients at the coarsest level are more stable and provide a crucial
signal when the images differ significantly, while the fine-scale gradients are
key to obtaining high spatial accuracy.
The loss we minimize is given by: 
\begin{equation}
    \mathbb{E}_{\RasterImage\sim\mathcal{D}}\sum_{l=1}^\NumPyrLevels \|\textit{pyr}_l(\RasterImage) - \RasterOutput_l\|^2,
    \label{eq:loss}
\end{equation}
where \NumPyrLevels is the number of pyramid levels,
$\textit{pyr}_l(\RasterImage)$ the $l$-th pyramid level, $\RasterOutput_l$ our
output rasterized at the corresponding spatial resolution, and $\mathcal{D}$ the
training dataset.

\subsection{Shape correspondences by segmentation}\label{sec:coloring}
When specializing a generative models to a single class, e.g., the same glyph or digit across multiple fonts, it is often desirable that the model's latent space capture correspondences between parts of the instance, like the opening in the capital letter `A', or the eyes and mouth of an emoji face.
To enable this, we segment our raster training dataset using an automatic off-the-shelf tool~\cite{kopf2011depixelizing}. We cluster these segments across the dataset based on spatial position, and assign to each cluster a unique RGB colour.
This consistent labeling helps learn a more interpretable latent space for purposes of interpolation, but is not itself critical; we show in supplementary material that our reconstruction is robust to inconsistent labeling thanks to the differentiable compositing step.
%

\subsection{Training details}
We train our model end-to-end for 100 -- 1000 epochs, using a batch size between
2 -- 256 and the Ranger optimizer~\cite{tong2019calibrating} with learning rate
between $10^{-3}$ and $10^{-4}$, depending on the dataset.
To evaluate path decoder's generalization to variable number of segments,
we randomly chose the number of segments $\NumSegments\in\{7,\ldots, 25\}$ at
every iteration.

\section{Evaluation}
We demonstrate \name's quantitative performance in 3 tasks: reconstruction,
generation, and interpolation. We compare it with raster based ImageVAE and
vector based SVG-VAE, DeepSVG on all the tasks.   

\begin{figure}[!tbh]
    \centering
    \includegraphics[width=\linewidth]{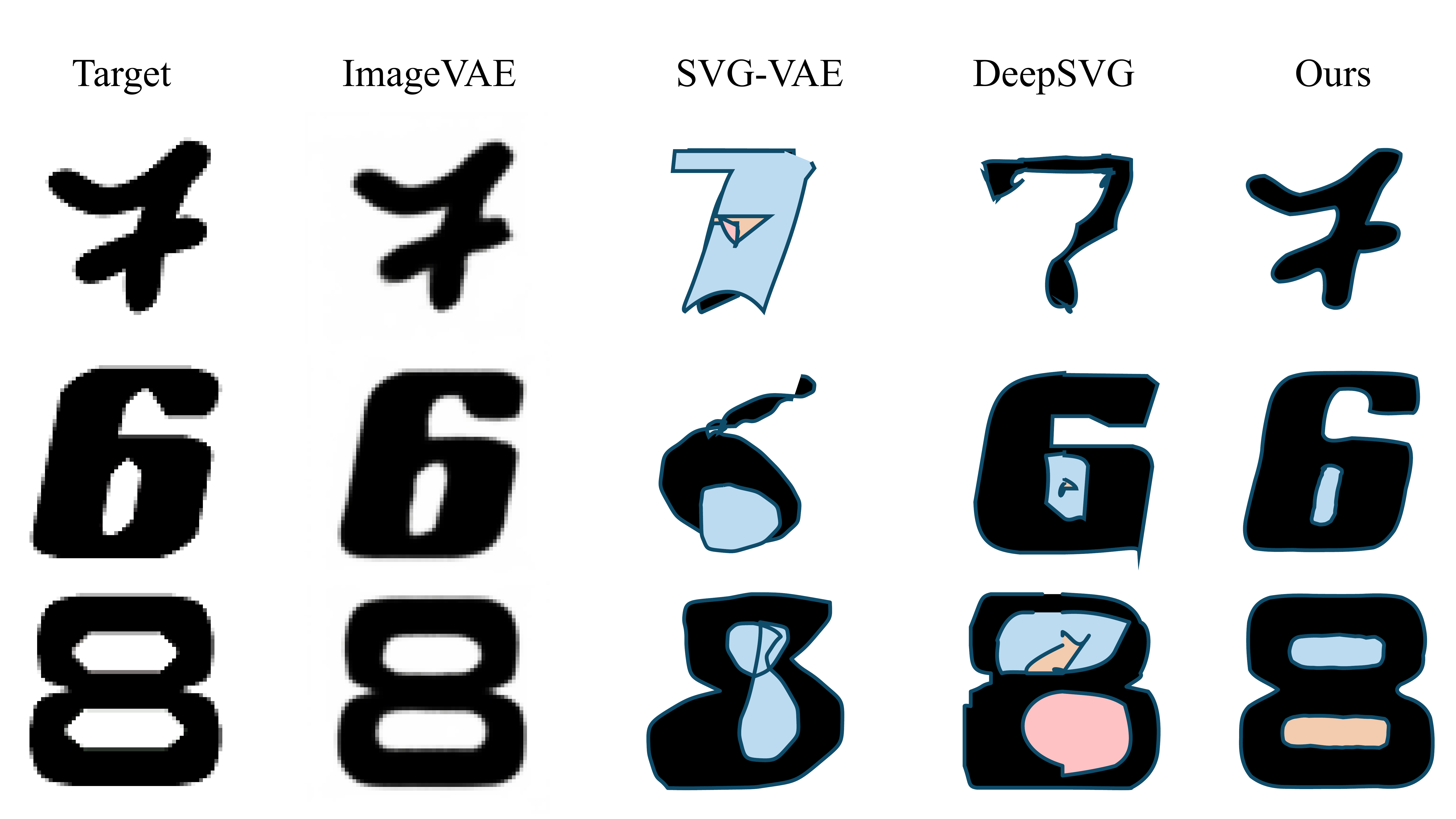}
    \caption{\label{fig:fonts_reconstruction_comparison}
        {\bf Reconstructions on \Fonts.}
        Our model, \name, captures complex topologies and produces vector outputs.
        ImageVAE has good fidelity but produces raster outputs with limited
        resolution (see Table~\ref{tbl:comparison}).
        SVG-VAE and DeepSVG produce vector outputs but often fail to accurately
        reproduce complex fonts.
        All the methods were trained on the same set of fonts.
        Please use digital zoom to better appreciate the quality of the vector graphics. 
    }
    
\end{figure}

\paragraph{Reconstruction} 
We measure the reconstruction performance of the baselines and Im2Vec using
$L_2$ loss in image space. This quantifies how accurately the latent space of
the different methods captures the training dataset. Since SVG-VAE and DeepSVG
work in vector domain, we rasterize their vector estimates using
CairoSVG~\cite{kozea}.

Table~\ref{tbl:comparison} shows reconstruction quality of the Im2Vec and other
baselines on \Fonts~\cite{lopes2019learned},
\MNIST~\cite{lecun-mnisthandwrittendigit-2010}, \Emojis~\cite{notoEmoji}, and
\Icons~\cite{creativeStall}. While vector based methods have the advantage of
being able reproduce the exact intended vector parametrization, they are
adversely effected  by the non-linear relationship between vector parameters and
image appearance. Therefore what seems like a small error in the vector
parameters estimated by SVG-VAE and DeepSVG may result in dramatic changes in
appearance. Unlike vector domain methods, \name is not affected by the objective
mismatch between the vector parameter and pixel spaces, thereby achieving
significant improvement in the reconstruction task. 

Refer to our supplementary for a chamfer distance based reconstruction comparison between SVG-VAE, DeepSVG and our method.

\begin{table}[t!]
  \centering
  \caption{\label{tbl:comparison}
    {\bf Reconstruction quality.} Comparison of pixel-space reconstruction losses for various methods
    and datasets. Note that neither SVG-VAE nor DeepSVG operate on datasets
    without vector supervision.
}
  \begin{tabular}{r c c c c}
    \toprule
  & \Fonts & \MNIST & \Emojis & \Icons \\ 
  \midrule
    ImageVAE & 0.0116 & 0.0033 & 0.0016 & 0.0002\\  
    SVG-VAE & 0.1322 & \ding{53} & - & -\\
    DeepSVG & 0.0938 & \ding{53} & - & -\\
    \name (Ours) & 0.0284 & 0.0036 & 0.0014 & 0.0003 \\
    \bottomrule
  \end{tabular}
\end{table}

\begin{figure}[t!]
    \centering
    \begin{subfigure}[b]{\linewidth}
        \centering
        \includegraphics[width=\linewidth]{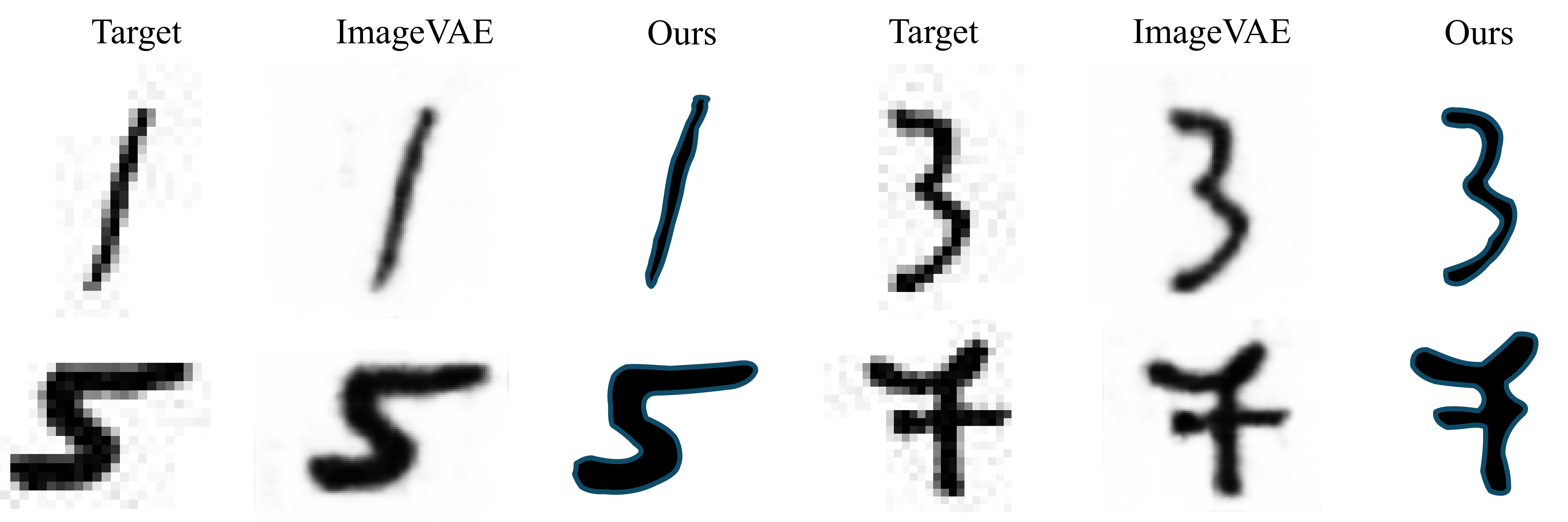}
        \caption{\label{fig:mnist_reconstruction}reconstruction}
    \end{subfigure}
    \begin{subfigure}[b]{\linewidth}
        \includegraphics[width=\linewidth]{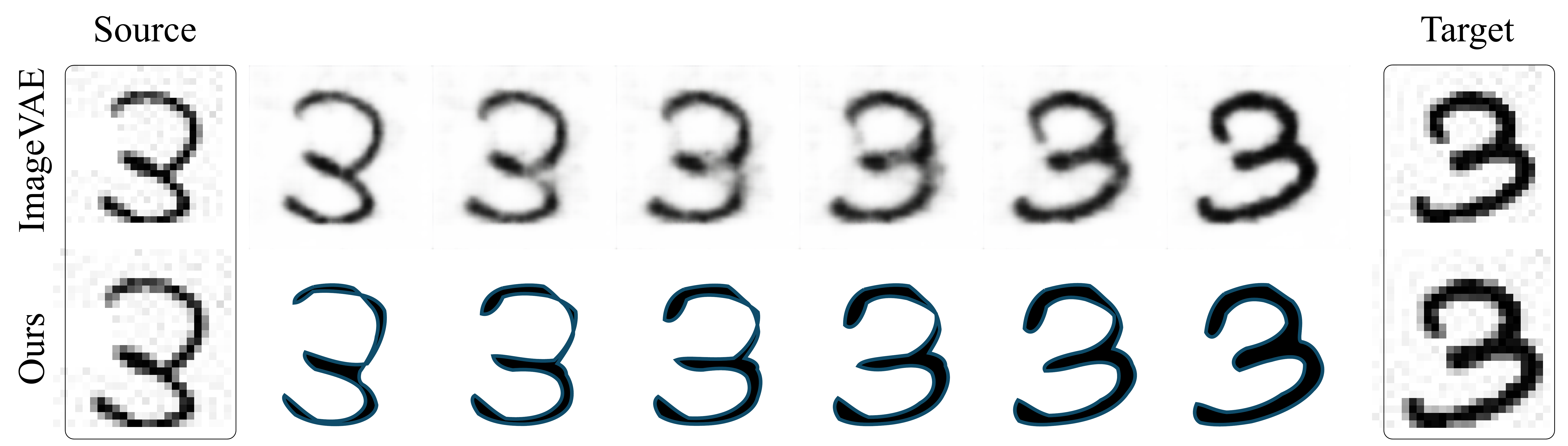}
        \caption{\label{fig:mnist_interpolation}interpolation}
    \end{subfigure}
    \caption{\label{fig:mnist_results}
        {\bf \MNIST results.}
        The \MNIST dataset only provides raster data.
        Since no vector graphics ground truth is available, neither SVG-VAE nor
        DeepSVG can be trained on this dataset. 
        We trained both ImageVAE and \name on the full dataset, with no
        digit class specialization or conditioning.
        Our model produces vector outputs, while ImageVAE is limited to
        low-resolution raster images ({\bf top}).
        Both models produce convincing interpolation ({\bf bottom}).
    }
\end{figure}

\begin{figure}[t!]
    \centering
    \includegraphics[width=\linewidth]{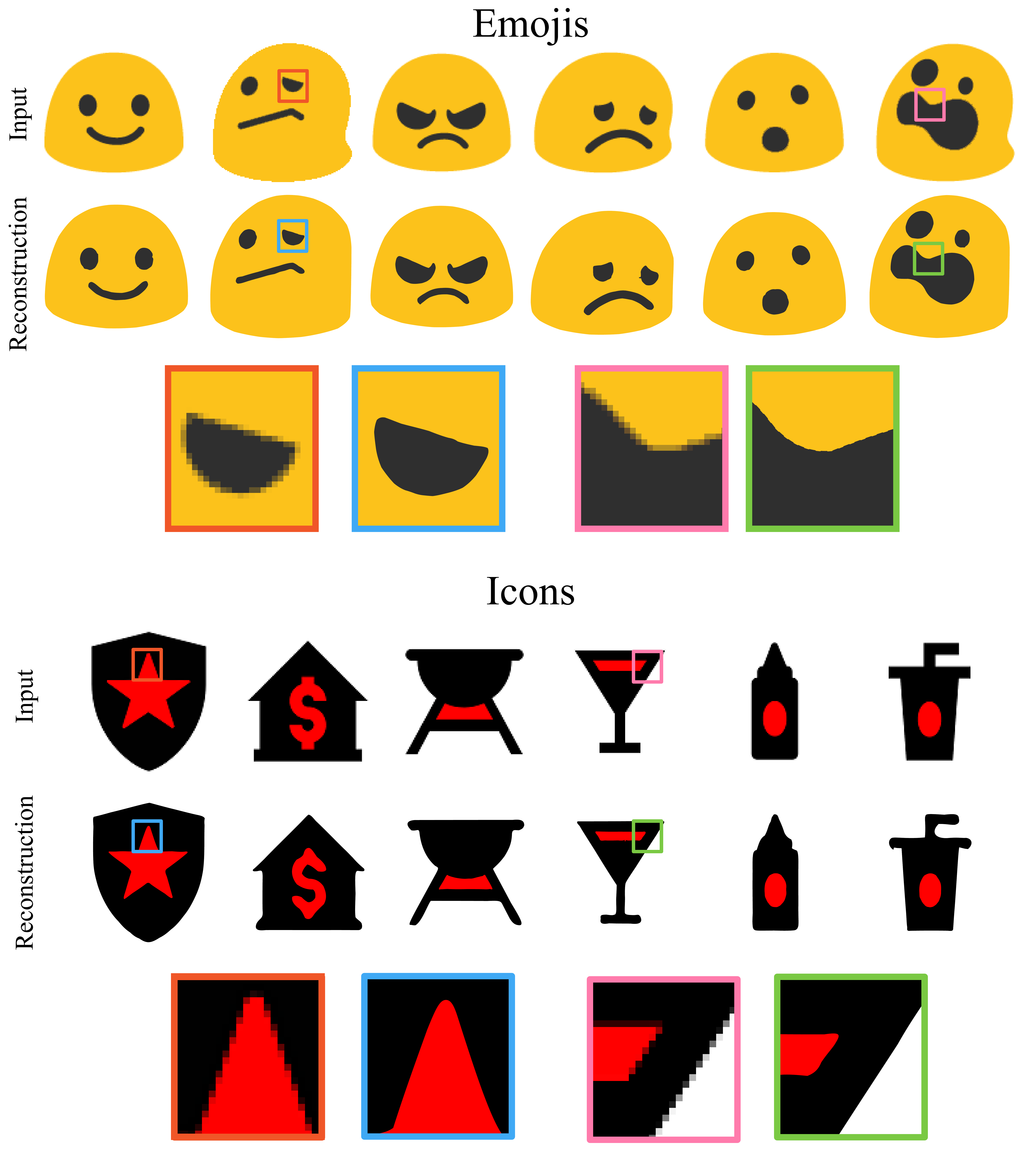}
    \caption{\label{fig:emoji_icons_reconstruction}
        {\bf Reconstructions.}
        Results on the \Emojis and the \Icons datasets.
        In each case, we show the input image ($128\times128$) and the corresponding
        vector graphics output, which can be rasterized at arbitrary resolutions. 
    }
    
\end{figure}

We show qualitative comparisons of input shape reconstruction between methods in
Figures~\ref{fig:fonts_reconstruction_comparison}
and~\ref{fig:mnist_reconstruction}. We also show reconstruction output of Im2Vec
on \Emojis and \Icons in Fig.~\ref{fig:emoji_icons_reconstruction}.

\begin{figure*}[t!]
    \centering
    \begin{subfigure}[b]{.49\linewidth}
        \includegraphics[width=\linewidth]{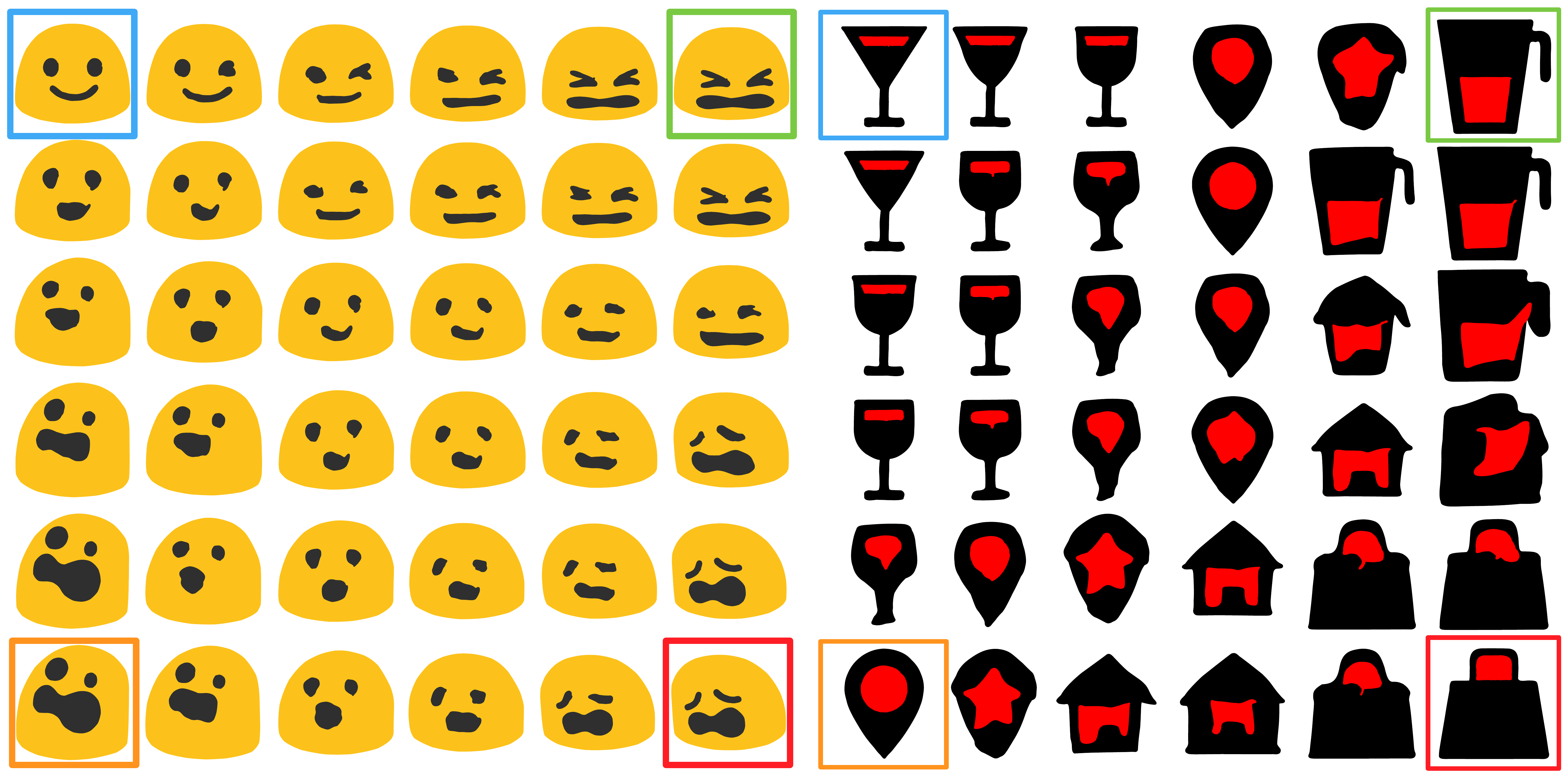}
        \caption{\label{fig:emoji_icons_Interpolation}\Emojis and \Icons
        interpolations using \name}
    \end{subfigure}
    \hfill
    \hfill
    \begin{subfigure}[b]{.47\linewidth}
        \includegraphics[width=\linewidth]{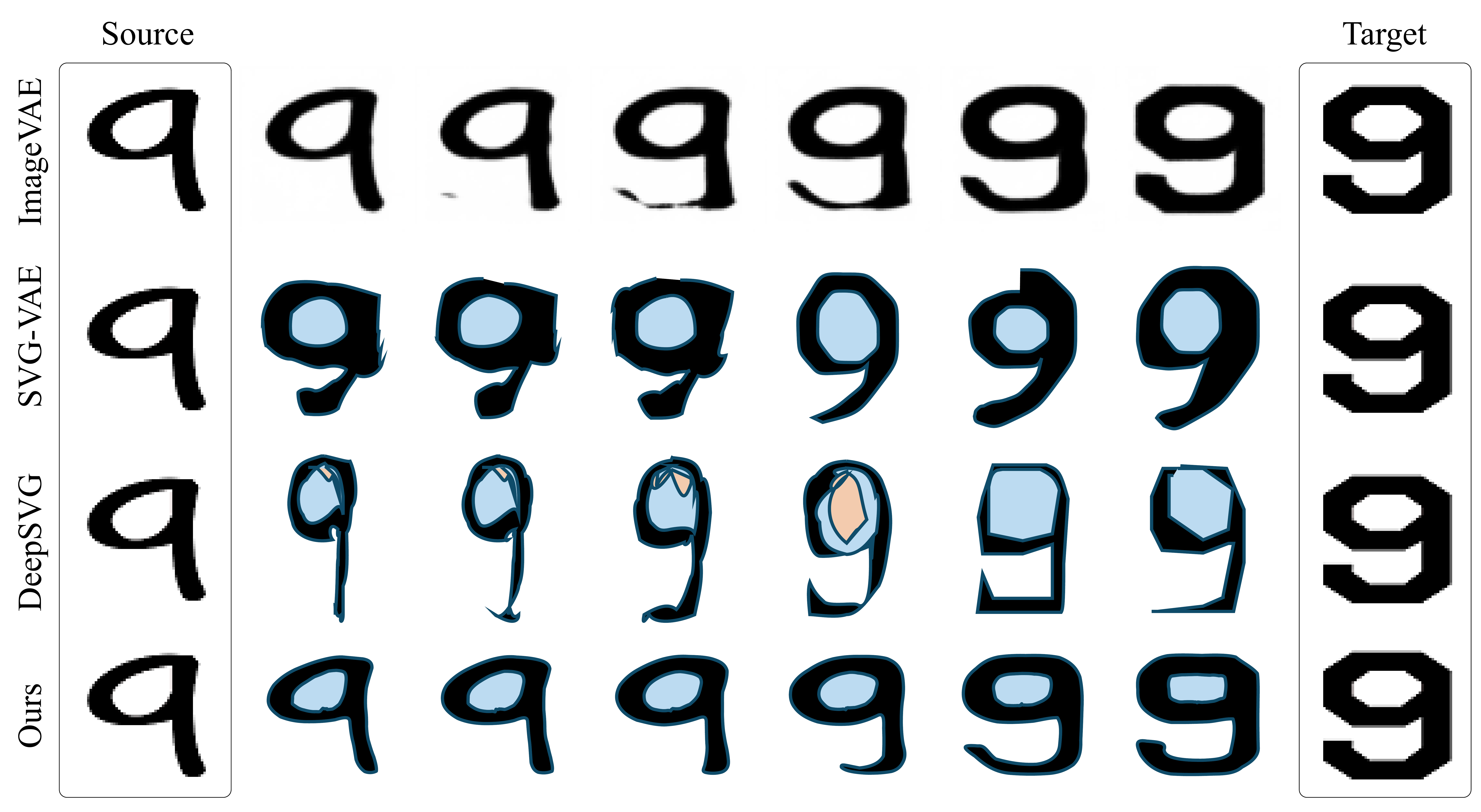}
        \caption{\label{fig:font_interpolation_comparison}Comparison to
        baselines on \Fonts}
    \end{subfigure}
    \caption{\label{fig:interpolations}
        {\bf Interpolations.}
        Our learned latent space enables plausible interpolation between samples.
        In~\subref{fig:emoji_icons_Interpolation}, we show interpolations between
        source--target pairs on the \Emojis and \Icons datasets.
        In~\subref{fig:font_interpolation_comparison} we show interpolations on
        the \Fonts dataset.
        Unlike previous work, \name enables plausible interpolation even across
        significant changes in shape.
        For instance, the stem of the digit `9'
        naturally curls along the interpolation path. 
    }
\end{figure*}

\begin{figure}[t!]
    \centering
    \includegraphics[width=\linewidth]{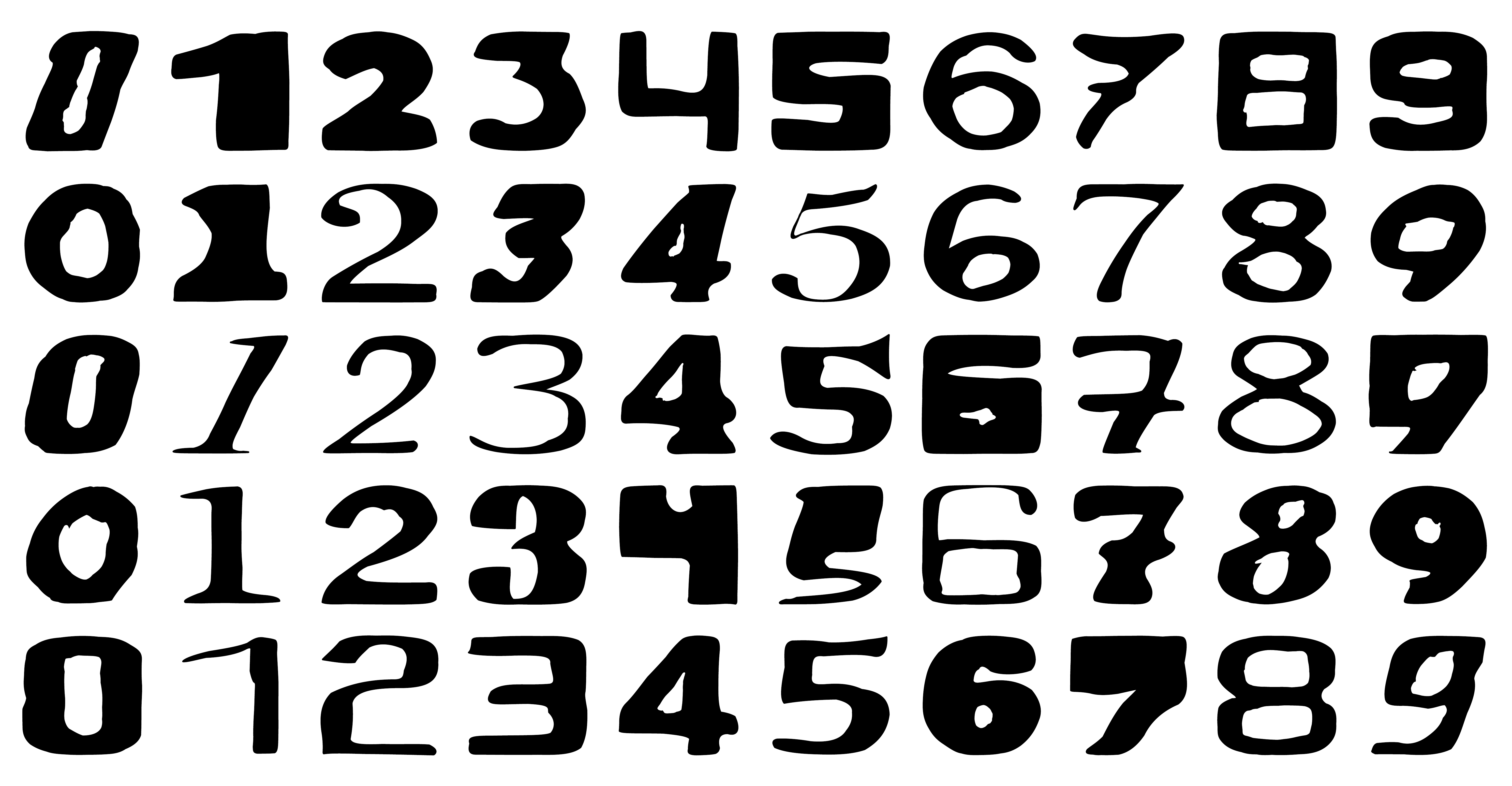}
    \caption{
        {\bf Random samples.}
        We show a random selection of digits generated by \name. The latent
        space was trained on the full Fonts dataset.
        Our model is capable of generating samples with significant topological
        variations across the different font types.
        In the supplemental material, we include 1000 random samples from the latent space.
        Please use digital zoom to better evaluate the quality. 
    }
    \label{fig:ours generation}
\end{figure}

\paragraph{Generation and Interpolation}
We present a random sample of font glyphs generated using Im2Vec in
Figure~\ref{fig:ours generation}. A qualitative comparison of latent space
interpolation between baselines and Im2Vec is presented in
Figures~\ref{fig:font_interpolation_comparison}
and~\ref{fig:mnist_interpolation}. We also present latent space interpolation
between 4 input images of \Emojis and \Icons in
Fig.~\ref{fig:emoji_icons_Interpolation}.

\begin{table}[h!]
  \centering
  \caption{\label{tab:generation}
    {\bf Generation and Interpolation quality.}
      Results on the \Fonts and
      the \MNIST are more accurate than both
      previous techniques that require vector supervision, and an image-based
      baseline autoencoder.
  }
  \begin{tabular}{r  c c c c}
    \toprule
    & \multicolumn{2}{c}{\bf Generation} & \multicolumn{2}{c}{\bf Interpolation} \\ 
  & \Fonts & \MNIST & \Fonts & \MNIST \\ 
  \midrule 
    ImageVAE & 0.171&  0.058 & 0.184 & 0.072 \\  
    SVG VAE & 0.206& \ding{53} & 0.206& \ding{53} \\
    DeepSVG & 0.210& \ding{53} & 0.202& \ding{53} \\
    \name (Ours) & 0.187 &  0.069 & 0.188 & 0.0872  \\
    \bottomrule
  \end{tabular} 
\end{table}
%

%
To quantitatively evaluate our generation results with others, we quantify how
realistic the intermediate shapes in the latent shape as the average closest
distance between the intermediate shapes to any sample in the training dataset:
\begin{equation}
\sum_{\RasterOutput \in \RasterOutput_G} \operatorname{min}_{I \in \mathrm{dataset}}(\|I, \RasterOutput\|^2), 
\end{equation}
where  $\RasterOutput_G$ is the set of all generated shapes. We variationally sample
1000 shapes from all the methods and present the quality of the generated shapes
in Table~\ref{tab:generation}. 

We perform similar evaluation to quantify the quality of our interpolations. For
comparison we sample 4 evenly spaced interpolations between 250 random pairs of
images from the training dataset to create interpolation samples. The results of
the quality of interpolation between different methods is presented in
Table~\ref{tab:generation}. 

\section{Limitations}
\begin{figure}[t]
    \centering
    \includegraphics[width=\linewidth]{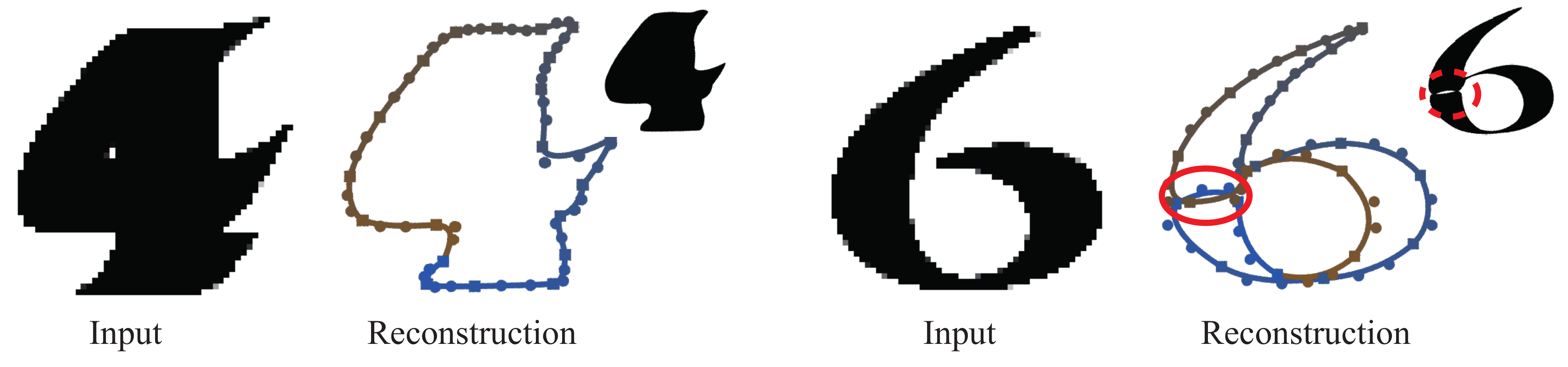}
    \caption{
        {\bf Limitations.}
        \name is only supervised by an image-space loss, so it can sometimes
        miss small topological features ({\bf Left}), or produce semantically
        meaningless or degenerate geometries ({\bf Right}).
        While the former can be resolved by providing higher resolution
        supervision, the later could be mitigated by using local geometric
        priors. 
    }
    \label{fig:failure_cases}
\end{figure}

The raster-based nature of the training imposes the principal limitations of our method (see Figure~\ref{fig:failure_cases}). It is possible for some very fine features to underflow the training resolution, in which case they may be lost. This could be addressed by increasing the resolution at the expense of computational efficiency, or perhaps by developing a more involved image-space loss.
Secondly, in particularly difficult cases it is possible for the generated shape to go to a local optimum that contains degenerate features or semantically non-meaningful parts which nonetheless still result in a plausible rasterised image. This is a consequence of lack of vector supervision, but could possibly be addressed by imposing geometric constraints on the generated paths.

\section{Conclusion}
We presented \name as a generative network that can be trained to produce vector graphics output of varying complexity and topology using only image supervision, without requiring vector sequence guidance. Our generative setup supports projection (i.e., converting images to vector sequences), sampling (i.e., generating new shape variations directly in vector form), as well as interpolation (i.e., morphing from one vector sequence to another, even with topological variations). Our evaluations show that \name achieves better reconstruction fidelity compared to methods requiring vector supervision.

We hope that this method can become the fundamental building block for neural processing of vector graphics and similar parametric shapes.


{\small
\bibliographystyle{ieee}
\bibliography{vector_vae}
}
\newcommand{\beginsupplement}{%
        \setcounter{section}{0}
        \renewcommand{\thesection}{S.\arabic{section}}%
        \setcounter{table}{0}
        \renewcommand{\thetable}{S\arabic{table}}%
        \setcounter{figure}{0}
        \renewcommand{\thefigure}{S\arabic{figure}}%
     }


\end{document}